\documentclass{article}
\usepackage{arxiv}

\usepackage[utf8]{inputenc}
\usepackage[T1]{fontenc}
\usepackage{hyperref}
\usepackage{url}
\usepackage{booktabs}
\usepackage{amsfonts}
\usepackage{amsmath}
\usepackage{amssymb}
\usepackage{tabularx}
\usepackage{array}
\usepackage{graphicx}
\usepackage{multirow}
\usepackage{float}
\usepackage{nicefrac}
\usepackage{microtype}
\usepackage[numbers,sort&compress]{natbib}
\usepackage{xcolor}
\usepackage{pifont}

\newcolumntype{C}{>{\centering\arraybackslash}X}

\hypersetup{
  colorlinks=true,
  linkcolor=blue!60!black,
  citecolor=blue!60!black,
  urlcolor=blue!60!black,
}

\title{IndustrialVLA-Bench: A Traceable Multi-Axis Evaluation\\ of Open Robot Policy Models}
\renewcommand{\shorttitle}{IndustrialVLA-Bench}

\author{%
  \normalfont\normalsize
  \parbox{\dimexpr\textwidth-2\tabcolsep\relax}{%
    \centering
    {\bfseries
      Yiqi Wang\textsuperscript{1,\,*}\quad
      Zhifeng Rao\textsuperscript{2,\,*}\quad
      Jiaqi Zhang\textsuperscript{3}\quad
      Xiaoyang Li\textsuperscript{4}\quad
      Zhangkai Wu\textsuperscript{5}\quad
      Yiqun Duan\textsuperscript{6}\\[0.35em]
      Mingkai Zheng\textsuperscript{7,\,\dag}\quad
      Fei Wang\textsuperscript{8}\quad
      Shan You\textsuperscript{9}\quad
      Taotao Cai\textsuperscript{10}%
    }\\[0.9em]
    \small
    \textsuperscript{1}Griffith University, Queensland, Australia \quad
    \textsuperscript{2}Southern University of Science and Technology, Shenzhen, China\\
    \textsuperscript{3}Jiangsu University, Zhenjiang, China \quad
    \textsuperscript{4}University of Southern Queensland, Brisbane, Australia\\
    \textsuperscript{5}The University of Sydney, Sydney, Australia \quad
    \textsuperscript{6}Norve Labs Inc., California, United States\\
    \textsuperscript{7}Southern University of Science and Technology, Shenzhen, China \quad
    \textsuperscript{8}ACE Robotics, Shanghai, China\\
    \textsuperscript{9}ACE Robotics, Shanghai, China \quad
    \textsuperscript{10}University of Southern Queensland, Brisbane, Australia\\[0.7em]
    \textsuperscript{*}Equal Contribution: \texttt{yiqi.wang.jennie@gmail.com}\quad
    \textsuperscript{\dag}Corresponding author: \texttt{zhengmk@sustech.edu.cn}%
  }%
}

\date{}

\begin{document}
\maketitle

\begin{abstract}
Open robot policies increasingly follow two paradigms: vision-language-action models (VLAs) directly map observations and instructions to actions, whereas world-action models (WAMs) incorporate learned video or world dynamics into policy learning or action generation. Although both target the same manipulation tasks and represent alternative design choices, they are commonly
reported under different evaluation protocols, leaving their capability, robustness, language sensitivity, and deployment-cost trade-offs unclear.
We present \textbf{IndustrialVLA-Bench}, an evidence-aware evaluation of six released VLA and WAM systems under a unified reporting schema. It separately evaluates clean capability on LIBERO, non-language robustness on LIBERO-Plus, instruction sensitivity on LIBERO-Para, and observed execution cost. 
Reported task scores aggregate three complete evaluations with distinct random seeds under a fixed checkpoint and inference configuration. 
Across all six systems, clean LIBERO averages differ by only 1.58 points,
whereas robustness and paraphrase summaries span 14.62 and 31.08 points.
Restricting every comparison to the three protocol-faithful systems
preserves the effect (1.36, 14.62 and 23.10 points), so the diagnostic
separation reported here does not depend on the weaker evidence tiers.
We additionally report observed inference latency, peak memory, runtime mode, and an evidence status for every system. Protocol-faithful, near-reproduction, and pending-verification entries remain visibly separated; only protocol-faithful entries support strict comparisons. Rather than claiming universal superiority of either paradigm, IndustrialVLA-Bench provides traceable evidence for comparing released robot policies on shared practical criteria. Code and evaluation records are available at \url{https://github.com/xiaoqi-7/IndustrialVLA-Bench}.
\end{abstract}


\section{Introduction}
\label{sec:introduction}

\begin{figure}[t]
\centering
\includegraphics[width=0.86\linewidth]{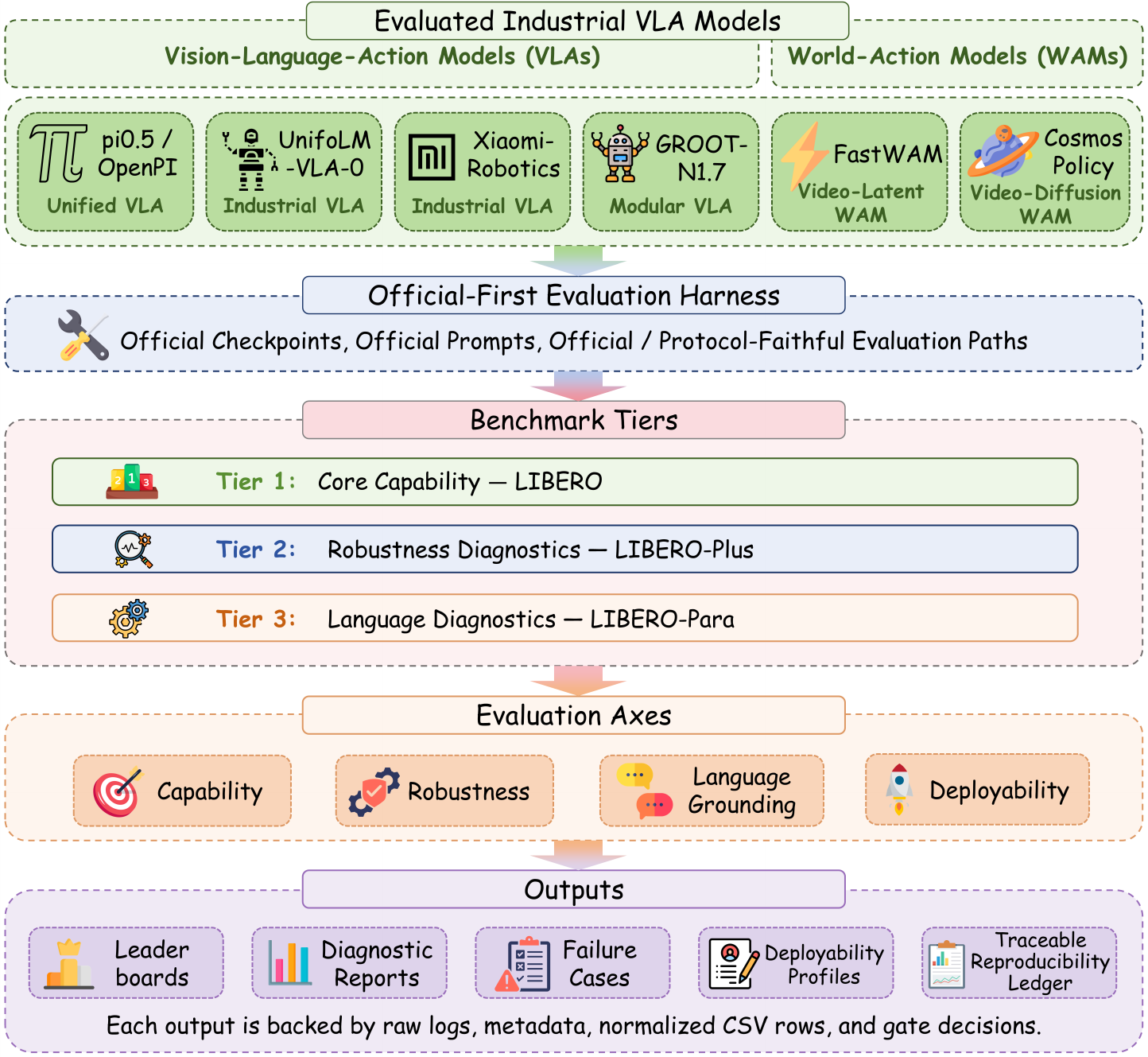}
\caption{Overview of IndustrialVLA-Bench. Six released VLA and World-Action Model systems are assessed on three complementary LIBERO-family tracks through four coordinated views: clean capability, robustness, language sensitivity, and execution and reproducibility evidence. Evidence status is kept explicit rather than hidden inside a single leaderboard.} %
\label{fig:unified-vla-view}
\end{figure}

Open robot policies increasingly follow two design paradigms.
\emph{Vision-language-action models} (VLAs) map visual observations and language instructions to actions, whereas \emph{world-action models} (WAMs) incorporate learned video or world dynamics into policy learning or action generation. Their internal mechanisms differ, but both target the same
instruction-conditioned manipulation tasks and therefore constitute alternative choices for research and deployment
\cite{black2025pi05,nvidia2026groot17,yuan2026fastwam,kim2026cosmospolicy}.
\textbf{Comparing them is practically necessary: users need to know not only whether a released policy succeeds on familiar tasks, but also whether it remains stable under visual and linguistic variation and whether it can be reproduced within available compute.} Representative foundations for open robot policies include RT-1, RT-2, 
Open X-Embodiment, Octo, and OpenVLA 
\cite{brohan2023rt1,zitkovich2023rt2,oneill2024openx,
ghosh2024octo,kim2025openvla}, while more recent VLA and WAM releases 
diversify policy and action-generation designs 
\cite{black2024pi0,cai2026xiaomi,kim2026cosmospolicy,
nvidia2026groot17,black2025pi05,unitree2026unifolmvla,
yuan2026fastwam}.

Current reporting does not support that decision reliably. First, even scores on the same benchmark may be generated from different checkpoints, prompts, observation-action interfaces, wrappers, trial counts, termination rules, and aggregation procedures. A numerical difference can therefore reflect either
policy capability or the evaluation path. Second, standard LIBERO success is close to saturation for several released systems, so a clean-task ranking can hide substantial differences under camera, embodiment, appearance, or instruction shifts. Third, latency, memory, runtime architecture, and
reproduction failures are usually separated from task performance, although they directly affect whether an open policy is usable.

Existing benchmarks and infrastructures address substantial parts of this problem. LIBERO evaluates language-conditioned manipulation \cite{liu2023libero}; LIBERO-Plus and LIBERO-Para diagnose non-language perturbations and semantics-preserving paraphrases \cite{fei2026liberoplus,kim2026liberopara}; VLA-Arena jointly varies task structure, language, and visual observations \cite{zhang2025vlaarena}; and MultiNet, VLA-Eval, and LeRobot provide complementary adaptation, evaluation, and deployment support \cite{guruprasad2025multinet,choi2026vlaeval,cadene2026lerobot}. Most importantly, a concurrent robustness study already compares prominent VLAs and WAMs on LIBERO-Plus and RoboTwin~2.0-Plus \cite{zhang2026wamrobustness}. IndustrialVLA-Bench therefore does not claim to be the first VLA--WAM robustness comparison. Its distinct contribution is to combine clean capability, non-language robustness, a separate paraphrase track, repeated run-level variation, observed execution cost, and explicit evidence-promotion states in one official-checkpoint evaluation.

We introduce \textbf{IndustrialVLA-Bench}\footnote{\url{https://github.com/xiaoqi-7/IndustrialVLA-Bench}}, a deliberately bounded evaluation of six publicly released systems. 
The name refers to the industrially released nature of the evaluated
systems, several of which are shipped by robotics vendors for deployment
use; it does not denote an industrial task domain. All task evidence in
this paper is simulation-based and LIBERO-family-specific.
The set covers unified and modular VLAs, a latent-video WAM, and a diffusion-based WAM, but is purposive rather than exhaustive. We report clean LIBERO, non-language perturbations from LIBERO-Plus, and instruction paraphrases from LIBERO-Para using three run-level seeds with a fixed checkpoint and inference configuration. An official-checkpoint evidence policy records repositories, checkpoints, commands, normalized outputs, and protocol deviations. Results that remain near-reproductions or pending verification are retained for coverage but are not promoted to strict rankings. 

The results illustrate the need for separate diagnostic axes. Within the protocol-faithful subset, clean LIBERO averages span only 1.36 points, whereas robustness and paraphrase scores span 14.62 and 23.10 points. NR/PV rows remain visible for coverage but do not enter strict comparisons. These findings describe the evaluated checkpoints rather than causal effects of VLA or WAM architectures.

We make the following contributions: 
\begin{itemize}
\item \textbf{Problem and protocol.} We formulate cross-paradigm robot-policy comparison as a multi-axis, evidence-constrained evaluation problem, with explicit comparability gates for official and protocol-faithful runs.
\item \textbf{Benchmark and artifact.} We provide a bounded six-model evaluation spanning released VLA and WAM designs, three complementary task tracks, repeated run-level evaluation, normalized result records, and explicit evidence states.
\item \textbf{Multi-axis evidence.} We jointly report clean capability, non-language robustness, instruction sensitivity, latency, memory, runtime mode, setup burden, and reproduction status without collapsing them into a single score.
\item \textbf{Empirical findings.} We show that near-saturated clean success obscures larger diagnostic differences and report checkpoint-level capability--robustness--cost profiles without attributing them causally to architecture.
\end{itemize}

\begin{table}[t]
\centering
\small
\setlength{\tabcolsep}{4.0pt}
\renewcommand{\arraystretch}{1.10}
\caption{
Positioning of IndustrialVLA-Bench against selected closely related robot-policy benchmarks and evaluation infrastructures.
\checkmark\ denotes explicit support or reporting,
$\triangle$ denotes partial or benchmark-dependent support,
and -- denotes that the dimension is not a primary focus.
}
\label{tab:related_infrastructure}

\begin{tabularx}{\textwidth}{
    l
    c
    c
    c
    c
    c
    c
    c
    >{\centering\arraybackslash}X
}
\toprule
\textbf{Framework}
& \textbf{\shortstack{Open\\source}}
& \textbf{\shortstack{Checkpoint\\reruns}}
& \textbf{\shortstack{VLA +\\WAM}}
& \textbf{\shortstack{Clean\\tasks}}
& \textbf{\shortstack{Visual\\shifts}}
& \textbf{\shortstack{Language\\shifts}}
& \textbf{\shortstack{Multi-seed\\reporting}}
& \textbf{\shortstack{Deployment and\\provenance reporting}} \\
\midrule

MultiNet
& \checkmark
& $\triangle$
& $\triangle$
& \checkmark
& $\triangle$
& --
& --
& Partial \\

VLA-Eval
& \checkmark
& \checkmark
& --
& \checkmark
& $\triangle$
& $\triangle$
& $\triangle$
& Evaluation infrastructure \\

LeRobot
& \checkmark
& $\triangle$
& $\triangle$
& $\triangle$
& --
& --
& --
& Training and deployment stack \\

\shortstack[l]{VLA-Arena\\\cite{zhang2025vlaarena}}
& \checkmark
& \checkmark
& --
& \checkmark
& \checkmark
& \checkmark
& --
& Task benchmark and toolchain \\

\shortstack[l]{WAM--VLA robustness\\study~\cite{zhang2026wamrobustness}}
& $\triangle$
& \checkmark
& \checkmark
& $\triangle$
& \checkmark
& \checkmark
& --
& Runtime; limited provenance audit \\

\textbf{IndustrialVLA-Bench}
& \checkmark
& \checkmark
& \checkmark
& \checkmark
& \checkmark
& \checkmark
& \checkmark
& \textbf{Latency, VRAM, runtime, setup, and provenance} \\

\bottomrule
\end{tabularx}

\vspace{0.4em}
\begin{minipage}{0.98\textwidth}
\footnotesize
\textbf{Notes.}
``Checkpoint reruns'' indicates that released model checkpoints are
independently executed rather than compared only through originally
reported results. ``Visual shifts'' and ``Language shifts'' denote
controlled robustness and paraphrase evaluation, respectively.
``Deployment and provenance reporting'' covers inference latency,
peak VRAM, runtime mode, setup burden, checkpoint provenance,
protocol deviations, and reproduction status.
\end{minipage}
\end{table}

\section{A Unified View of Open Robot-Policy Evaluation}
\label{sec:unified-view}

A central challenge in benchmarking open VLA systems is that their
architectures, action representations and runtime designs are
heterogeneous: released policies differ in whether actions are produced by
a single vision-language stack, a separate action module, a flow or
diffusion head, or a learned video representation~\cite{kim2025openvla,black2024pi0,black2025pi05,liu2025rdt,pertsch2025fast,cai2026xiaomi}. Comparing such systems through final success rates alone is misleading when the scores are produced under different architectural and execution assumptions.

\begin{figure}[!htbp]
    \centering
    \includegraphics[width=0.78\textwidth]{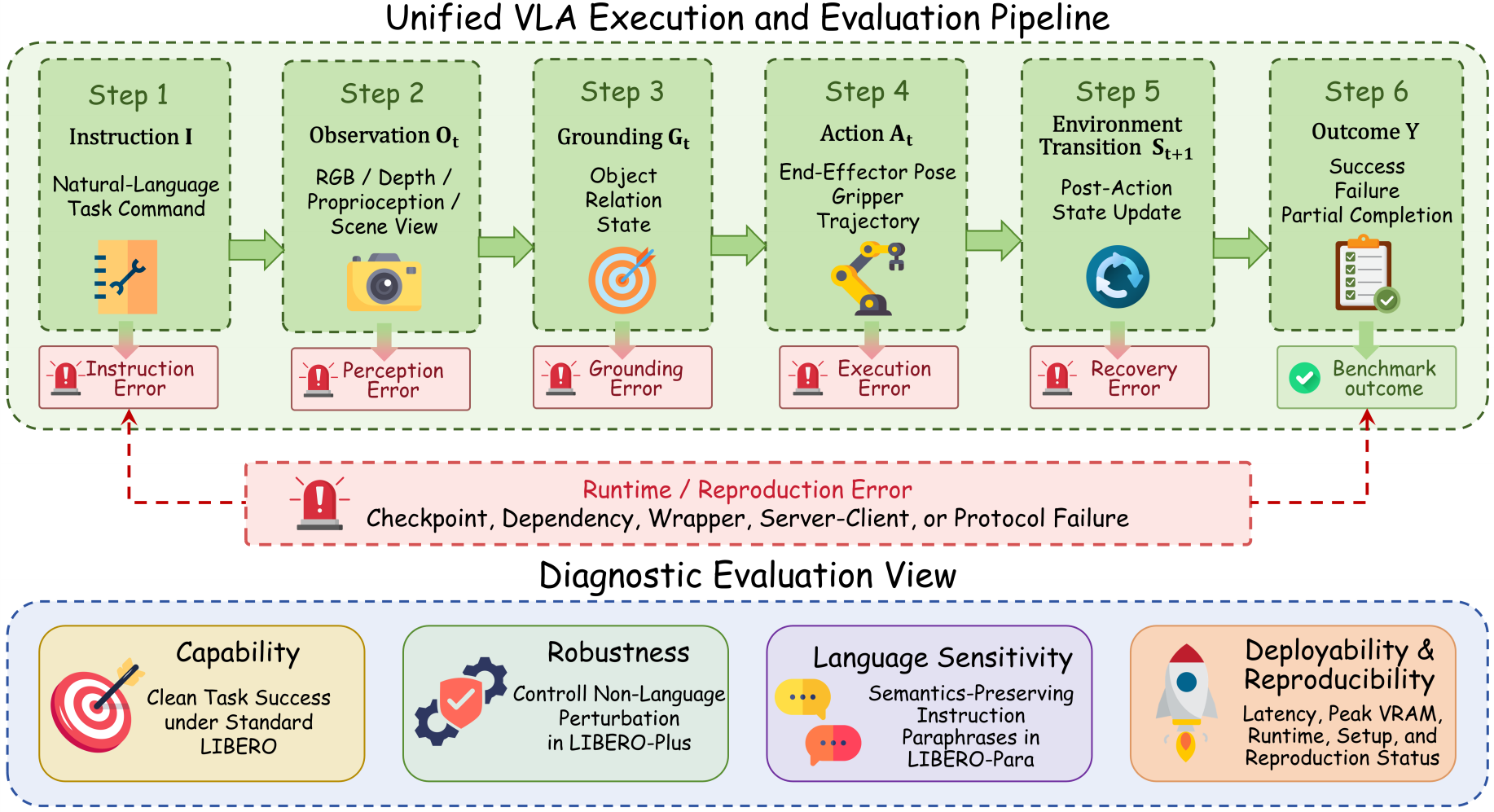}
    \caption{Unified execution and evaluation VLA and WAM systems. Although different models use heterogeneous architectures, action representations, and runtime designs, their evaluation behavior can be mapped to a shared instruction-conditioned execution pipeline. Each stage corresponds to diagnostic evaluation dimensions, including capability, robustness, language sensitivity, and deployability and reproducibility.}
    \label{fig:unified-vla-pipeline}
\end{figure}

We therefore map every evaluation episode onto a shared instruction-conditioned execution view: an instruction and an observation stream are grounded into task-relevant entities, processed by policy inference, translated into executable actions, and judged through the resulting environment transitions. Figure~\ref{fig:unified-vla-pipeline} illustrates this view and its correspondence to the diagnostic axes; each execution stage exposes a distinct source of failure.

The shared contract localizes where evidence is observed, but aggregate success rates do not identify why an episode failed. We therefore do not assign instruction, perception, planning, control, or recovery labels without episode-level annotation. Runtime failures, including unsupported checkpoints, dependency conflicts, timeouts, or action-normalization mismatches, are recorded separately as reproduction evidence rather than inferred from task scores.

This view yields four complementary axes. \textbf{Capability} is clean task success under the standard LIBERO protocol \cite{liu2023libero}. \textbf{Robustness} is performance degradation under controlled visual, camera, embodiment, and layout perturbations, instantiated by LIBERO-Plus. \textbf{Language sensitivity} is whether performance survives paraphrases that preserve task semantics, evaluated with LIBERO-Para \cite{kim2026liberopara}. \textbf{Deployability and reproducibility} cover inference latency, peak VRAM, runtime mode, setup burden, checkpoint provenance, and the strength of the reproduced evaluation path.

These axes are deliberately not collapsed into one scalar: a model with high clean success may be brittle under paraphrase, expensive to run, or hard to reproduce, so we report cross-axis model profiles rather than a clean-success-only ranking. Appendix~\ref{app:execution-view} formalizes the execution contract and evidence tiers.

\section{Benchmark Design and Protocol}
\label{sec:benchmark-design}

Rather than introducing a new architecture, IndustrialVLA-Bench defines an evaluation protocol that makes heterogeneous VLA releases comparable without hiding differences in benchmark support, runtime assumptions, or engineering cost.

\subsection{Positioning Against Existing Evaluation Infrastructure}
\label{sec:positioning}

MultiNet \cite{guruprasad2025multinet}, VLA-Eval \cite{choi2026vlaeval}, and LeRobot \cite{cadene2026lerobot} address complementary parts of the robot-policy lifecycle but differ from IndustrialVLA-Bench in their primary evaluation objective; Table~\ref{tab:related_infrastructure} summarizes the distinction.

IndustrialVLA-Bench is an evidence protocol rather than a general-purpose robot-learning library. It independently executes the named VLA and WAM checkpoints, reports three-run summaries, and separates clean capability, visual robustness, paraphrase sensitivity, runtime observations, and verification state. The following subsections define the scope and metrics; Appendix~\ref{app:evidence} formalizes the comparability gates.

\subsection{Model Scope}
\label{sec:modelscope}

IndustrialVLA-Bench evaluates six publicly released models: $\pi_{0.5}$\slash OpenPI, UnifoLM-VLA-0, Xiaomi-Robotics-0, GR00T-N1.7, FastWAM, and Cosmos Policy. The July 2026 snapshot applied five executability criteria covering checkpoint availability, instruction-conditioned manipulation support, LIBERO interface compatibility, documentable inference configuration, and independence from closed remote APIs (Appendix~\ref{app:evidence}).

The selected models form a deliberately bounded, coverage-driven case set rather than a census of open robot policies. $\pi_{0.5}$/OpenPI represents a \emph{single-stack} VLA,  whereas GR00T-N1.7 separates vision-language processing from an action-generation component. UnifoLM-VLA-0 and Xiaomi-Robotics-0 add independently released VLA systems with different scales and runtimes. FastWAM and Cosmos Policy extend the comparison to WAMs with different uses of future-video representations. These labels describe released implementations and are not controlled architectural variables. %

DreamZero is excluded because no publicly executable checkpoint was available at the snapshot. Other public systems are outside scope because this release prioritizes three-track coverage over exhaustive model enumeration. The reported scores for all six systems use the same three-seed aggregation, but their public execution evidence is not equally mature: OpenPI, GR00T-N1.7, and FastWAM are protocol-faithful; UnifoLM-VLA-0 is a near-reproduction; and Xiaomi-Robotics-0 and Cosmos Policy remain pending verification. Tables retain the latter entries for transparent diagnostic coverage and mark them explicitly.

\subsection{Benchmark Scope}
\label{sec:benchmarkscope}

IndustrialVLA-Bench focuses on four experimentally completed evaluation axes. Clean capability is measured on the four standard LIBERO suites: LIBERO-Spatial, LIBERO-Object, LIBERO-Goal, and LIBERO-Long. Robustness is evaluated using LIBERO-Plus under controlled changes to camera viewpoint, robot appearance or embodiment, illumination, background, visual noise, and object layout. Language sensitivity is evaluated using LIBERO-Para, which modifies instruction wording while preserving task semantics. Deployability is evaluated through the benchmark harness using peak VRAM, per-step inference latency, parameter count, diffusion configuration, runtime mode, setup burden, and reproduction metadata.

The current release is limited to these tracks and claims no experimental
coverage of other suites or of equal-budget adaptation. Appendix~\ref{app:evidence} provides the full scope statement and lists the excluded settings.

\subsection{Metrics}
\label{sec:metrics}

IndustrialVLA-Bench reports metrics along four complementary axes. \textbf{Capability} is the clean episode success rate under the standard LIBERO protocol,
\[
\mathrm{SR}(m,s)=\frac{1}{N_s}\sum_{i=1}^{N_s}\mathbb{I}\left[\mathrm{success}_i\right],
\]
where $N_s$ is the number of evaluation episodes for model $m$ in suite $s$; we report per-suite rates and their unweighted average over the four standard suites. \textbf{Robustness} is success under each controlled non-language perturbation in LIBERO-Plus, together with the absolute drop $\Delta_{\mathrm{robust}}=\mathrm{SR}_{\mathrm{clean}}-\mathrm{SR}_{\mathrm{pert}}$ and the corresponding retention ratio; camera, robot or embodiment, illumination, background, visual-noise, and object-layout perturbations are reported separately. The LIBERO-Plus language condition is shown for completeness but excluded from this average. \textbf{Language sensitivity} is paraphrase success on LIBERO-Para together with the clean-to-paraphrase drop $\Delta_{\mathrm{para}}=\mathrm{SR}_{\mathrm{clean}}-\mathrm{SR}_{\mathrm{para}}$, using the same clean-task checkpoint without adaptation. The two language tracks use different perturbation constructions and are not pooled. 
\textbf{Execution and reproduction evidence} comprises observed latency per policy call, peak VRAM, runtime mode, single-GPU feasibility, setup burden, and checkpoint provenance, and verification status. These fields are reported separately rather than compressed into a single scalar score.

\section{Experiments}
\label{sec:experiments}

We organize the evaluation around four questions: how well the evaluated systems solve standard LIBERO tasks under directly comparable conditions (RQ1); how much performance is retained when visual observations, camera configurations, robot appearance, and object layouts are perturbed in LIBERO-Plus (RQ2); how well task performance is preserved under semantics-preserving instruction paraphrases in LIBERO-Para (RQ3); and what computational and engineering costs reproduction and deployment require (RQ4). %

\subsection{Experimental Setup}
\label{sec:experimental-setup}

We evaluate the six systems of Section~\ref{sec:modelscope} on the three benchmark families of Section~\ref{sec:benchmarkscope}, using released checkpoints and official evaluation implementations wherever available.
When no official end-to-end path exists, a documented reproduction preserves task definitions, observation and action semantics, success conditions, episode horizon, and aggregation rules. Evidence status is attached to every row; a numerical result does not by itself establish protocol fidelity.

Every reported model--benchmark configuration aggregates three complete runs with fixed run-level seeds $\{1,7,42\}$, holding the checkpoint and inference configuration fixed. A run seed initializes the simulator reset and task-order generators; it is neither an episode index nor a training seed. Standard LIBERO uses 50 trials per task (2{,}000 episodes per seed, 6{,}000 over three seeds), while LIBERO-Plus and LIBERO-Para use one episode per official task configuration (10{,}030 and 4{,}092 episodes per seed). Tables report the run-level mean and the population standard deviation over the three runs. These deviations are descriptive, not confidence intervals or hypothesis tests; where mean gaps are comparable to run variation, we avoid a resolved rank claim.
Appendix~\ref{app:setup} gives the formulas and measurement caveats, %
and Appendix~\ref{sec:harness-reproducibility} details the evidence records. %

\subsection{Clean Capability on LIBERO}
\label{sec:rq1-clean-capability}

RQ1 evaluates the selected models on clean manipulation tasks under the standard LIBERO task definitions. LIBERO serves as the shared-denominator benchmark because all six systems map to its instruction-conditioned observation--action interface with limited semantic adaptation.

\begin{table}[t]
\centering
\small
\setlength{\tabcolsep}{4.5pt}
\renewcommand{\arraystretch}{1.10}
\caption{Clean LIBERO evaluation summary. Each value is the mean success rate over three run-level seeds. Superscripts report public evidence status: PF = protocol-faithful, NR = near-reproduction, and PV = pending verification. Only PF rows are eligible for strict comparisons. %
}
\label{tab:libero_main_leaderboard}

\begin{tabular}{lcccccl}
\toprule
\textbf{Model}
& \textbf{Spatial}
& \textbf{Object}
& \textbf{Goal}
& \textbf{Long}
& \textbf{Average}
& \textbf{Trials} \\
\midrule

\multicolumn{7}{c}{\textbf{Vision-Language-Action Models (VLAs)}} \\
\cmidrule(lr){1-7}

$\pi_{0.5}$ / OpenPI\textsuperscript{PF}
& 98.33 $\pm$ 0.09
& 98.67 $\pm$ 0.50
& 97.93 $\pm$ 0.09
& 91.40 $\pm$ 0.65
& 96.52 $\pm$ 0.22
& 6{,}000 (50/task) \\

UnifoLM-VLA-0\textsuperscript{NR}
& 98.67 $\pm$ 0.66
& 100.00 $\pm$ 0.00
& 97.73 $\pm$ 0.25
& 95.27 $\pm$ 0.41
& 97.92 $\pm$ 0.06
& 6{,}000 (50/task) \\

Xiaomi-Robotics-0\textsuperscript{PV}
& 98.67 $\pm$ 0.09
& 100.00 $\pm$ 0.00
& 97.93 $\pm$ 0.94
& 95.80 $\pm$ 0.91
& 98.10 $\pm$ 0.43
& 6{,}000 (50/task) \\

GR00T-N1.7\textsuperscript{PF}
& 97.60 $\pm$ 1.50
& 99.30 $\pm$ 0.50
& 99.30 $\pm$ 0.50
& 95.30 $\pm$ 1.00
& 97.88 $\pm$ 0.65
& 6{,}000 (50/task) \\

\midrule
\multicolumn{7}{c}{\textbf{World-Action Models (WAMs)}} \\
\cmidrule(lr){1-7}

FastWAM \textsuperscript{PF}
& 97.07 $\pm$ 0.25
& 99.13 $\pm$ 0.09
& 96.47 $\pm$ 0.50
& 93.53 $\pm$ 0.41
& 96.55 $\pm$ 0.16
& 6{,}000 (50/task) \\

Cosmos Policy\textsuperscript{PV}
& 96.40 $\pm$ 0.33
& 99.60 $\pm$ 0.33
& 97.93 $\pm$ 0.34
& 96.60 $\pm$ 0.28
& 97.63 $\pm$ 0.23
& 6{,}000 (50/task) \\

\bottomrule
\end{tabular}

\vspace{0.35em}
\begin{minipage}{0.97\textwidth}
\footnotesize
\textbf{Note.} Average is the unweighted mean of the four suite-level success rates. NR and PV rows are retained for coverage but must not be read as evidence-equivalent to PF rows.%
\end{minipage}
\end{table}

\textbf{Overall clean performance.} Table~\ref{tab:libero_main_leaderboard} shows that all six
models achieve high clean-task performance, with average success rates
between 96.52\% and 98.10\%. Among protocol-faithful systems the clean
averages are GR00T-N1.7 97.88\%, FastWAM 96.55\% and OpenPI 96.52\%. The
highest numerical entries, Xiaomi-Robotics-0 at 98.10\% (PV) and
UnifoLM-VLA-0 at 97.92\% (NR), are not rank-eligible under our evidence
policy and are listed for coverage only. Several adjacent mean gaps are
smaller than at least one associated run-level standard deviation, so the
present $n=3$ summaries do not support a statistically resolved total
ordering.

\paragraph{LIBERO-Long remains the most discriminative clean suite.} Performance is close to saturation on LIBERO-Spatial and LIBERO-Object, where most models exceed 97\%, whereas LIBERO-Long produces the lowest success rate for five of the six models --- most visibly for OpenPI and FastWAM at 91.40\% and 93.53\%. Aggregate success rates alone cannot determine whether this gap arises from planning, control, or recovery.

\paragraph{Clean success alone provides limited separation.} The full reported range is only 1.58 percentage points. This motivates the diagnostic tracks, but it does not imply that all systems are equivalent or that small clean differences are significant.

\subsection{Robustness under LIBERO-Plus}
\label{sec:rq2-robustness}

RQ2 evaluates whether models preserve task success when the underlying task semantics remain unchanged but the visual and embodiment conditions differ from the clean evaluation setting. We use six LIBERO-Plus perturbation dimensions: camera viewpoint, robot appearance or embodiment, illumination, background, visual noise, and object layout. Language perturbations are excluded from this subsection and analyzed separately in Section~\ref{sec:rq3-language}.

\begin{table}[t]
\centering
\small
\setlength{\tabcolsep}{3.2pt}
\renewcommand{\arraystretch}{1.10}
\caption{LIBERO-Plus diagnostic results over three run-level seeds. Robust Avg. excludes the Language column and averages the six non-linguistic dimensions. Status superscripts follow Table~\ref{tab:libero_main_leaderboard}.}%
\label{tab:libero_plus_robustness}
\resizebox{\textwidth}{!}{%
\begin{tabular}{lccccccccc}
\toprule
\textbf{Model} & \textbf{Camera} & \textbf{Robot} & \textbf{Language} &
\textbf{Light} & \textbf{Background} & \textbf{Noise} & \textbf{Layout} &
\textbf{Robust Avg.} & \textbf{Drop}  \\
\midrule
\multicolumn{10}{c}{\textbf{Vision-Language-Action Models (VLAs)}} \\
\cmidrule(lr){1-10}
$\pi_{0.5}$ / OpenPI\textsuperscript{PF}
  & 70.63 $\pm$ 0.37 & 75.12 $\pm$ 0.76 & 85.97 $\pm$ 0.08
  & 96.76 $\pm$ 0.47 & 95.82 $\pm$ 0.42 & 87.05 $\pm$ 0.36
  & 86.47 $\pm$ 0.26 & 85.31 $\pm$ 0.32 & 11.21 $\pm$ 0.38
  \\
UnifoLM-VLA-0\textsuperscript{NR}
  & 57.03 $\pm$ 0.31 & 68.37 $\pm$ 0.87 & 91.09 $\pm$ 0.80
  & 93.70 $\pm$ 0.25 & 95.14 $\pm$ 0.19 & 79.26 $\pm$ 0.45
  & 79.19 $\pm$ 0.56 & 78.78 $\pm$ 0.30 & 19.14 $\pm$ 0.36
   \\
Xiaomi-Robotics-0\textsuperscript{PV}
  & 40.22 $\pm$ 0.09 & 55.63 $\pm$ 0.20 & 89.02 $\pm$ 0.24
  & 94.51 $\pm$ 0.43 & 90.43 $\pm$ 0.55 & 86.78 $\pm$ 0.11
  & 75.87 $\pm$ 0.18 & 73.91 $\pm$ 0.13 & 24.19 $\pm$ 0.33
  \\
GR00T-N1.7\textsuperscript{PF}
  & 64.40 $\pm$ 0.89 & 38.40 $\pm$ 1.57 & 83.88 $\pm$ 1.02
  & 94.87 $\pm$ 0.95 & 93.72 $\pm$ 0.19 & 84.36 $\pm$ 0.25
  & 74.94 $\pm$ 0.69 & 75.12 $\pm$ 0.28 & 22.76 $\pm$ 0.28
  \\
\midrule
\multicolumn{10}{c}{\textbf{World-Action Models (WAMs)}} \\
\cmidrule(lr){1-10}
FastWAM\textsuperscript{PF}
  & 44.65 $\pm$ 1.54 & 72.04 $\pm$ 0.35 & 66.99 $\pm$ 1.28
  & 94.22 $\pm$ 1.12 & 66.57 $\pm$ 0.56 & 67.17 $\pm$ 0.82
  & 79.45 $\pm$ 0.32 & 70.69 $\pm$ 0.52 & 25.86 $\pm$ 0.42
   \\
Cosmos Policy\textsuperscript{PV}
  & 72.65 $\pm$ 0.58 & 52.56 $\pm$ 0.64 & 89.07 $\pm$ 0.14
  & 98.48 $\pm$ 0.36 & 84.94 $\pm$ 0.46 & 90.21 $\pm$ 0.12
  & 84.00 $\pm$ 0.30 & 80.47 $\pm$ 0.23 & 17.16 $\pm$ 0.01
   \\
\bottomrule
\end{tabular}}

\vspace{0.35em}
\begin{minipage}{0.98\textwidth}
\footnotesize
\textbf{Note.} $^\dagger$The LIBERO-Plus language condition is reported for completeness but excluded from Robust Avg. and Drop. LIBERO-Para (Table~\ref{tab:libero_para}) is the primary language-sensitivity track; the two language protocols are not pooled. NR/PV rows are provisional diagnostic evidence. %
\end{minipage}
\end{table}

\paragraph{Robustness separates profiles more clearly.} OpenPI obtains the
highest robustness average at 85.31\%, despite ranking below several models
on clean LIBERO. Cosmos Policy records 80.47\% and UnifoLM-VLA-0 records
78.78\%, but both carry non-protocol-faithful status and are reported as
provisional diagnostic evidence. Among protocol-faithful systems the
robustness ordering is OpenPI 85.31\%, GR00T-N1.7 75.12\% and FastWAM
70.69\%. In contrast, Xiaomi-Robotics-0 falls from the highest clean average
to 73.91\% under perturbations. These results show that clean-task rank is
not a reliable proxy for robustness.

\paragraph{Camera and robot shifts are the dominant failure modes.} Averaged across the six models, camera and robot perturbations achieve only 58.26\% and 60.35\% success, respectively. Illumination is much less damaging, with a cross-model average of 95.42\%. Five models reach their lowest perturbation-specific result under either camera or robot changes. The most severe cases are Xiaomi-Robotics-0 under camera changes at 40.22\% and GR00T-N1.7 under robot changes at 38.40\%. Appearance-level changes such as illumination are largely absorbed by the visual representation, whereas camera and robot shifts alter the mapping between image evidence and action-relevant geometry.

\paragraph{OpenPI provides the strongest balanced robustness profile.} OpenPI does not obtain the best score on every individual perturbation, but it avoids catastrophic degradation across all six dimensions. Its lowest result is 70.63\% under camera changes, and it has the smallest clean-to-robustness drop of 11.21 percentage points. This suggests that robustness depends not only on peak performance, but also on whether a model avoids a single dominant failure mode.

\paragraph{WAMs do not share a uniform robustness profile.} The two WAM rows differ substantially across camera, background, noise, and layout conditions. Because the checkpoints also differ in data, scale, inference, and evidence status, these observations do not identify an effect of WAM architecture. %

\subsection{Language Sensitivity under LIBERO-Para}
\label{sec:rq3-language}

RQ3 isolates instruction sensitivity from visual robustness. LIBERO-Para modifies the surface form of the language instruction while preserving the underlying task and environment. Models are evaluated without fine-tuning on the paraphrased test instructions; the same clean-task checkpoint is used for both standard LIBERO and LIBERO-Para evaluation.

\begin{table}[t]
\centering
\small
\setlength{\tabcolsep}{4.5pt}
\renewcommand{\arraystretch}{1.10}
\caption{Language sensitivity on LIBERO-Para over three run-level seeds. Status superscripts follow Table~\ref{tab:libero_main_leaderboard}.}
\label{tab:libero_para}
\begin{tabular}{lcc}
\toprule
\textbf{Model} & \textbf{Mean $\pm$ Std (\%)}  & \textbf{Clean-to-Para drop (pp)}  \\
\midrule
\multicolumn{3}{c}{\textbf{Vision-Language-Action Models (VLAs)}} \\
\cmidrule(lr){1-3}
$\pi_{0.5}$ / OpenPI\textsuperscript{PF}
  & 71.33 $\pm$ 0.05 & 25.19 $\pm$ 0.12\\
UnifoLM-VLA-0\textsuperscript{NR}
  & 82.24 $\pm$ 0.48  & 15.67 $\pm$ 0.53
  \\
Xiaomi-Robotics-0\textsuperscript{PV}
  & 75.73 $\pm$ 0.16  & 22.37 $\pm$ 0.48
   \\
GR00T-N1.7\textsuperscript{PF}
  & 74.26 $\pm$ 1.52  & 23.62 $\pm$ 1.52
   \\
\midrule
\multicolumn{3}{c}{\textbf{World-Action Models (WAMs)}} \\
\cmidrule(lr){1-3}
FastWAM\textsuperscript{PF}
  & 51.16 $\pm$ 0.38  & 45.39 $\pm$ 0.32
   \\
Cosmos Policy\textsuperscript{PV}
  & 70.50 $\pm$ 0.26  & 27.13 $\pm$ 0.18
   \\
\bottomrule
\end{tabular}

\vspace{0.35em}
\begin{minipage}{0.97\columnwidth}
\textbf{Note.} Drop is relative to clean LIBERO. No model is adapted or fine-tuned on LIBERO-Para. NR/PV rows are provisional diagnostic evidence.
\end{minipage}
\end{table}

\paragraph{Paraphrases separate the models far more than clean tasks.} LIBERO-Para success ranges from 51.16\% to 82.24\%, a spread of 31.08 percentage points. The same six systems differ by only 1.58 percentage points on clean LIBERO. This diagnostic contrast is large, but evidence status remains part of every comparison. 

\paragraph{The lowest paraphrase degradation is recorded by a near-reproduction entry.} 
UnifoLM-VLA-0 records 82.24\% and loses only 15.67 points relative to its
clean average, the smallest degradation we observe; because this row is NR,
we report it as diagnostic coverage rather than as a rank claim. Among
protocol-faithful systems, GR00T-N1.7 retains the most (74.26\%), followed
by OpenPI (71.33\%) and FastWAM (51.16\%).
Because no model is adapted to the paraphrased instructions, these differences reflect how strongly each released checkpoint depends on the surface form of its training instructions.

\paragraph{FastWAM exhibits a language-grounding bottleneck.} FastWAM falls to 51.16\%, a 45.39-point reduction from its clean average and the largest degradation we observe on any axis. Its paraphrase success is lower than its aggregate robustness average of 70.69\% and lower than five of its six individual perturbation scores; only camera shifts damage it more. Conditioning action prediction on video latents therefore does not by itself confer invariance to instruction wording, which suggests that visual prediction and instruction grounding remain separable capabilities.

\paragraph{Language robustness and visual robustness are not equivalent.} The two diagnostic axes rank the models differently. OpenPI obtains the highest LIBERO-Plus average at 85.31\% but only the fourth-highest LIBERO-Para score, whereas UnifoLM-VLA-0 leads on LIBERO-Para despite ranking third on LIBERO-Plus. Cosmos Policy shows a similar mismatch in the opposite direction, ranking second on robustness but fifth on paraphrase success. Visual invariance and instruction grounding therefore expose different weaknesses and should not be collapsed into a single robustness score.

The two language protocols disagree, which is itself a comparability
result. LIBERO-Plus contains a language perturbation condition (Table 3)
and LIBERO-Para is a dedicated paraphrase suite (Table 4). Both are
presented as measurements of instruction sensitivity, and we deliberately
do not pool them. Comparing them post hoc shows why. Their model orderings
agree only moderately (Spearman rho = 0.60), and the disagreement is
concentrated in individual systems: Cosmos Policy ranks second of six on
the LIBERO-Plus language condition (89.07) but fifth on LIBERO-Para
(70.50), while GR00T-N1.7 moves in the opposite direction, from fifth
(83.88) to third (74.26). The two protocols also differ in absolute
difficulty by 13.47 points on average (84.34 against 70.87), and that
offset is not uniform across models, ranging from 8.85 points for
UnifoLM-VLA-0 to 18.57 points for Cosmos Policy. A single "language
robustness" number would therefore depend substantially on which
perturbation construction produced it. This supports reporting the two
tracks separately rather than merging them, and it is an instance of the
general principle that comparability must be earned rather than assumed.

\subsection{Deployability and Reproducibility}
\label{sec:rq4-deployability}

RQ4 reports observed deployment profiles rather than controlled efficiency rankings. GR00T-N1.7 records the lowest observed latency and peak VRAM, whereas OpenPI combines the highest protocol-faithful robustness with the largest observed latency. FastWAM records the highest memory use, while the PV Cosmos Policy row records lower observed memory and latency.

These values may reflect hardware, precision, caching, preprocessing, communication, action chunking, and runtime implementation in addition to model design. We therefore do not infer an intrinsic capability-efficiency trade-off. Setup burden likewise remains an ordinal summary of installation, checkpoint preparation, wrapper requirements, runtime services, and hardware constraints.

\begin{table}[t]
\centering
\small
\setlength{\tabcolsep}{4.3pt}
\renewcommand{\arraystretch}{1.10}
\caption{Deployability and reproducibility summary. Peak VRAM is reported in MiB and latency is average wall-clock model inference time per policy call. All values correspond to the evaluated checkpoint and benchmark runtime.}
\label{tab:deployability_reproducibility}

\begin{tabularx}{\textwidth}{
    l
    ccc
    >{\centering\arraybackslash}X
    >{\centering\arraybackslash}X
    cc
}
\toprule
\textbf{Model}
& \textbf{\shortstack{Peak\\VRAM}}
& \textbf{\shortstack{Latency\\(ms/call)}}
& \textbf{Parameters}
& \textbf{\shortstack{Diffusion\\steps}}
& \textbf{\shortstack{Runtime\\mode}}
& \textbf{\shortstack{Setup\\burden}} 
& \textbf{Status} \\
\midrule

\multicolumn{8}{c}{\textbf{Vision-Language-Action Models (VLAs)}} \\
\cmidrule(lr){1-8}

$\pi_{0.5}$ / OpenPI
& 14,550 & 58.80 & 3.6B & 10
& Local / server-client & Medium--High & PF  \\

UnifoLM-VLA-0
& 18,332 & 19.82 & 8.9B & 4
& Local / model-specific & High & NR  \\

Xiaomi-Robotics-0
& 10,294 & 25.44 & 4.7B & 5
& Model-specific & High & PV \\

GR00T-N1.7
& 7,810 & 19.13 & 3.4B & 4
& Deployment-oriented & High & PF  \\

\midrule
\multicolumn{8}{c}{\textbf{World-Action Models (WAMs)}} \\
\cmidrule(lr){1-8}

FastWAM
& 26,902 & 45.33 & 12.4B & 10
& Deployment-oriented & High & PF  \\

Cosmos Policy
& 8,654 & 21.26 & 2.0B & 5
& Deployment-oriented & High & PV \\

\bottomrule
\end{tabularx}

\vspace{0.5em}
\begin{flushleft}
\footnotesize
\textbf{Note.} Peak VRAM is measured during evaluation-time inference after runtime warm-up. Latency excludes environment reset time. Setup burden summarizes dependency installation, checkpoint preparation, environment setup, wrapper requirements, runtime services, and hardware constraints. Reproduction status follows the evidence policy defined in Section~\ref{sec:experimental-setup}.
\end{flushleft}
\end{table}

\begin{figure}[t]
\centering
\includegraphics[width=0.86\textwidth]{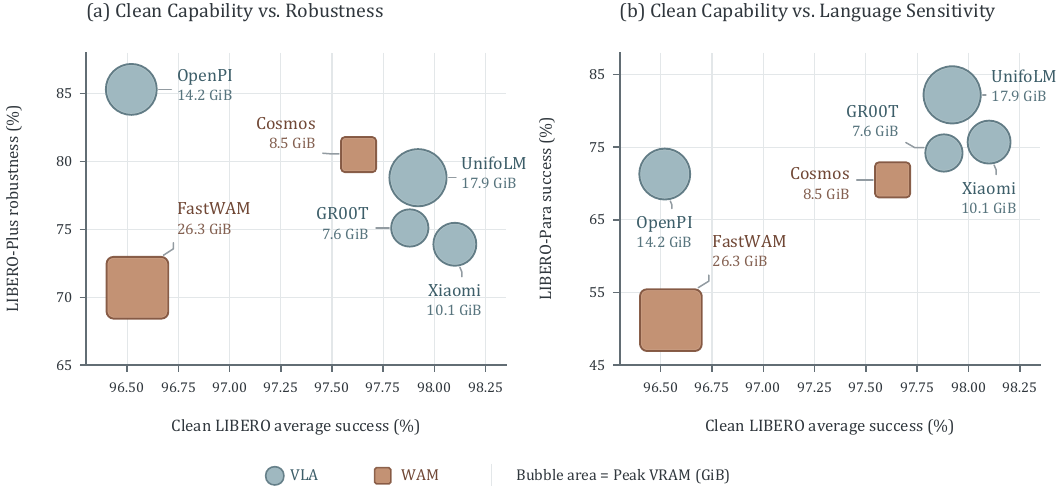}
\caption{Cross-axis comparison of the six evaluated models. The x-axis shows average clean LIBERO success; the y-axes show (a) LIBERO-Plus robustness over six non-linguistic perturbations and (b) LIBERO-Para language sensitivity. Bubble area denotes peak evaluation-time VRAM; marker shape distinguishes VLAs from WAMs.}
\label{fig:cross_axis_profiles}
\end{figure}

\subsection{Cross-Model Findings}
\label{sec:cross-model-findings}

Figure~\ref{fig:cross_axis_profiles} shows that clean capability, robustness, language sensitivity, and deployment cost are not aligned, and four patterns emerge across the axes.
All four patterns below are stated over the full six-system set for
coverage; each also holds when restricted to the three protocol-faithful
systems, with clean, robustness and paraphrase ranges of 1.36, 14.62 and
23.10 points respectively.
First, clean LIBERO is close to saturation: all six models fall within a 1.58-point range of 96.52\%--98.10\%, so clean success remains a useful capability check but provides limited discrimination. Second, clean ranking does not predict robustness: Xiaomi-Robotics-0 ranks first on clean LIBERO yet falls below OpenPI and Cosmos Policy on the six-dimension LIBERO-Plus average, and different perturbation families expose model-specific weaknesses rather than a single shared failure mode. Third, language sensitivity is partially decoupled from clean capability: FastWAM matches the others on clean LIBERO but drops to 51.16\% on LIBERO-Para, differing from UnifoLM-VLA-0 by 31.08 points there against only 1.37 points on clean tasks. Fourth, no evaluated model dominates every axis, and the WAM label alone does not confer robustness: OpenPI provides the strongest robustness but the highest latency, UnifoLM-VLA-0 the strongest language robustness but high VRAM, GR00T-N1.7 low latency and memory but marked sensitivity to robot perturbations, and Cosmos Policy a low observed-cost profile under pending verification (PV). These results support reporting model profiles across the four axes rather than selecting models through a single aggregate leaderboard.

\subsection{The diagnostic axes are not redundant}
Because we report the axes
separately, we can measure directly how much each one adds. Over the six
evaluated checkpoints, the Spearman rank correlation between clean LIBERO
average and LIBERO-Plus robustness average is -0.37, between clean average
and LIBERO-Para success is 0.77, and between LIBERO-Plus robustness and
LIBERO-Para success is 0.09. Clean success therefore does not merely fail
to predict the robustness ordering; over this checkpoint set the two
orderings run mildly opposite. The near-zero correlation between the visual
and the language diagnostic confirms that the two are not interchangeable
measurements of a single robustness property, and that collapsing them into
one score would discard most of the information each contributes. With n=6
these coefficients are descriptive rather than statistically significant,
but they quantify the separation that Figure 3 shows only qualitatively.

\section{Discussion and Limitations}
\label{sec:discussion}

\paragraph{Beyond clean leaderboards.} Clean task success is necessary but not sufficient for evaluating open robot policies: it does not reveal whether a model stays reliable under controlled visual and language variation, or whether it can be reproduced and deployed. This matters because VLA outputs are executable actions, so small errors compound over time, and because standard suites contain repeated layouts and limited instruction templates that reward visual priors and task shortcuts. For the same reason we do not collapse our axes into a single score: no objective weighting can decide how clean success should trade off against robustness retention, paraphrase sensitivity, latency, memory, or reproduction difficulty, and an aggregate would hide exactly the trade-offs our results expose. Users who want an out-of-the-box system should therefore prioritize setup burden and the reproduction evidence retained by the harness (Appendix~\ref{sec:harness-reproducibility}); users studying robustness should examine perturbation-specific rates and clean-to-perturbation drops; users targeting deployment should focus on latency, peak VRAM, and single-GPU feasibility.

\paragraph{Limitations.} The six-checkpoint set is purposive and non-exhaustive. All task results are simulation-based within the LIBERO family \cite{liu2023libero}; they do not validate industrial deployment, real-world sensing and contact, safety, or long-term reliability. Only three rows currently satisfy the protocol-faithful public-evidence state; one is a near-reproduction and two await verification. %
The n=3 deviations describe replay variation on a fixed configuration set,
not sampling uncertainty over task configurations and not training
uncertainty; no paired significance test is claimed, and we avoid resolved
rank claims wherever adjacent means fall within one run deviation.
Runtime measurements are not controlled for hardware, precision, batch, action-chunk length, or percentile latency. Setup burden is an ordinal audit rather than a timed user study. %

\paragraph{Ethics and responsible use.} The benchmark uses simulated manipulation tasks and no human--subject or identity data. Nevertheless, robot-policy scores can be misread as safety certification. Our results must not be used to justify unsupervised physical deployment: they omit collision severity, human proximity, hardware limits, and recovery under real-world failures. Physical deployment therefore requires separate risk assessment, controlled testing, and human oversight. Public checkpoints and code also retain their original licenses and use restrictions.

\paragraph{Artifact availability.} The evaluation harness, command manifests, normalized result records, and per-run logs are released at the repository linked above. The evaluated checkpoints remain under their original licences and are not redistributed. Upon acceptance we will deposit a versioned, DOI-bearing archival snapshot of the release and apply for the ACM Artifacts Available badge. New model entries are accepted through the promotion checklist of Appendix C.4 and are versioned against a pinned benchmark commit.

\section{Conclusion}
\label{sec:conclusion}

We introduced \textbf{IndustrialVLA-Bench}, a traceable multi-axis benchmark for comparing six publicly executable VLA and WAM checkpoints on three LIBERO-family tracks. The benchmark combines aligned repeated execution, explicit comparability gates, three complementary LIBERO-family tracks, deployment measurements, and traceable run records. Clean performance differs by only $1.58$ points, while non-language robustness and instruction sensitivity separate the evaluated systems much more strongly; no model dominates every axis. These findings show why VLA and WAM releases should be compared through evidence-bearing profiles rather than isolated clean-task scores, without implying universal superiority of either paradigm. 

\bibliographystyle{unsrtnat}
\bibliography{references} 

@inproceedings{brohan2023rt1,
  title     = {{RT-1}: Robotics Transformer for Real-World Control at Scale},
  author    = {Brohan, Anthony and Brown, Noah and Carbajal, Justice and Chebotar, Yevgen and Dabis, Joseph and Finn, Chelsea and Gopalakrishnan, Keerthana and Hausman, Karol and Herzog, Alex and Hsu, Jasmine and Ibarz, Julian and Ichter, Brian and Irpan, Alex and Jackson, Tomas and Jesmonth, Sally and Joshi, Nikhil J. and Julian, Ryan and Kalashnikov, Dmitry and Kuang, Yuheng and Leal, Isabel and Lee, Kuang-Huei and Levine, Sergey and Lu, Yao and Malla, Utsav and Manjunath, Deeksha and Mordatch, Igor and Nachum, Ofir and Parada, Carolina and Peralta, Jodilyn and Perez, Emily and Pertsch, Karl and Quiambao, Jornell and Rao, Kanishka and Ryoo, Michael and Salazar, Grecia and Sanketi, Pannag and Sayed, Kevin and Singh, Jaspiar and Sontakke, Sumedh and Stone, Austin and Tan, Clayton and Tran, Huong and Vanhoucke, Vincent and Vuong, Quan and Xia, Fei and Xiao, Ted and Xu, Peng and Xu, Sichun and Yu, Tianhe and Zitkovich, Brianna},
  booktitle = {Proceedings of Robotics: Science and Systems},
  year      = {2023},
  address   = {Daegu, Republic of Korea},
  month     = {July},
  url       = {https://arxiv.org/abs/2212.06817}
}

@inproceedings{zitkovich2023rt2,
  title     = {{RT-2}: Vision-Language-Action Models Transfer Web Knowledge to Robotic Control},
  author    = {Zitkovich, Brianna and Yu, Tianhe and Xu, Sichun and Xu, Peng and Xiao, Ted and Xia, Fei and others},
  booktitle = {Proceedings of the 7th Conference on Robot Learning},
  series    = {Proceedings of Machine Learning Research},
  volume    = {229},
  pages     = {2165--2183},
  publisher = {PMLR},
  year      = {2023},
  url       = {https://proceedings.mlr.press/v229/zitkovich23a.html}
}

@inproceedings{oneill2024openx,
  title     = {Open {X}-Embodiment: Robotic Learning Datasets and {RT-X} Models},
  author    = {{Open X-Embodiment Collaboration} and O'Neill, Abby and Rehman, Abdul and Gupta, Abhinav and Maddukuri, Abhiram and Gupta, Abhishek and Padalkar, Abhishek and Lee, Abraham and Pooley, Acorn and Gupta, Agrim and others},
  booktitle = {2024 IEEE International Conference on Robotics and Automation (ICRA)},
  pages     = {6892--6903},
  year      = {2024},
  doi       = {10.1109/ICRA57147.2024.10611477}
}

@inproceedings{ghosh2024octo,
  author    = {Ghosh, Dibya and Walke, Homer Rich and Pertsch, Karl and Black, Kevin and Mees, Oier and Dasari, Sudeep and Hejna, Joey and Kreiman, Tobias and Xu, Charles and Luo, Jianlan and Tan, You Liang and Chen, Lawrence Yunliang and Vuong, Quan and Xiao, Ted and Sanketi, Pannag R. and Sadigh, Dorsa and Finn, Chelsea and Levine, Sergey},
  title     = {{Octo}: An Open-Source Generalist Robot Policy},
  booktitle = {Proceedings of Robotics: Science and Systems},
  year      = {2024},
  address   = {Delft, Netherlands},
  month     = jul,
  doi       = {10.15607/RSS.2024.XX.090}
}

@inproceedings{kim2025openvla,
  title     = {{OpenVLA}: An Open-Source Vision-Language-Action Model},
  author    = {Kim, Moo Jin and Pertsch, Karl and Karamcheti, Siddharth and Xiao, Ted and Balakrishna, Ashwin and Nair, Suraj and Rafailov, Rafael and Foster, Ethan and Lam, Grace and Sanketi, Pannag and Vuong, Quan and Kollar, Thomas and Burchfiel, Benjamin and Tedrake, Russ and Sadigh, Dorsa and Levine, Sergey and Liang, Percy and Finn, Chelsea},
  booktitle = {Proceedings of the Conference on Robot Learning},
  series    = {Proceedings of Machine Learning Research},
  volume    = {270},
  year      = {2025},
  url       = {https://proceedings.mlr.press/v270/kim25c.html},
  eprint    = {2406.09246},
  archivePrefix = {arXiv}
}

@misc{black2024pi0,
  title         = {{$\pi_0$}: A Vision-Language-Action Flow Model for General Robot Control},
  author        = {Black, Kevin and Brown, Noah and Driess, Danny and Esmail, Adnan and Equi, Michael and Finn, Chelsea and Fusai, Niccolo and Groom, Lachy and Hausman, Karol and Ichter, Brian and Jakubczak, Szymon and Jones, Tim and Ke, Liyiming and Levine, Sergey and Li-Bell, Adrian and Mothukuri, Mohith and Nair, Suraj and Pertsch, Karl and Shi, Lucy Xiaoyang and Tanner, James and Vuong, Quan and Walling, Anna and Wang, Haohuan and Zhilinsky, Ury},
  year          = {2024},
  eprint        = {2410.24164},
  archivePrefix = {arXiv},
  primaryClass  = {cs.RO},
  url           = {https://arxiv.org/abs/2410.24164}
}

@inproceedings{black2025pi05,
  title     = {{$\pi_{0.5}$}: A Vision-Language-Action Model with Open-World Generalization},
  author    = {{Physical Intelligence} and Black, Kevin and Brown, Noah and Darpinian, James and Dhabalia, Karan and Driess, Danny and Esmail, Adnan and Equi, Michael and Finn, Chelsea and Fusai, Niccolo and Galliker, Manuel Y. and Ghosh, Dibya and Groom, Lachy and Hausman, Karol and Ichter, Brian and Jakubczak, Szymon and Jones, Tim and Ke, Liyiming and LeBlanc, Devin and Levine, Sergey and Li-Bell, Adrian and Mothukuri, Mohith and Nair, Suraj and Pertsch, Karl and Ren, Allen Z. and Shi, Lucy Xiaoyang and Smith, Laura and Springenberg, Jost Tobias and Stachowicz, Kyle and Tanner, James and Vuong, Quan and Walke, Homer and Walling, Anna and Wang, Haohuan and Yu, Lili and Zhilinsky, Ury},
  booktitle = {Proceedings of the Conference on Robot Learning},
  series    = {Proceedings of Machine Learning Research},
  volume    = {305},
  year      = {2025},
  url       = {https://proceedings.mlr.press/v305/black25a.html},
  eprint    = {2504.16054},
  archivePrefix = {arXiv}
}

@misc{unitree2026unifolmvla,
  author       = {{Unitree}},
  title        = {{UnifoLM-VLA-0}: A Vision-Language-Action (VLA) Framework under {UnifoLM} Family},
  year         = {2026},
  howpublished = {GitHub repository and model release},
  url          = {https://github.com/unitreerobotics/unifolm-vla},
  note         = {Accessed July 5, 2026}
}

@misc{cai2026xiaomi,
  title         = {Xiaomi-Robotics-0: An Open-Sourced Vision-Language-Action Model with Real-Time Execution},
  author        = {Cai, Rui and Guo, Jun and He, Xinze and Jin, Piaopiao and Li, Jie and Lin, Bingxuan and Liu, Futeng and Liu, Wei and Ma, Fei and Ma, Kun and Qiu, Feng and Qu, Heng and Su, Yifei and Sun, Qiao and Wang, Dong and Wang, Donghao and Wang, Yunhong and Wu, Rujie and Xiang, Diyun and Yang, Yu and Ye, Hangjun and Zhang, Yuan and Zhou, Quanyun},
  year          = {2026},
  eprint        = {2602.12684},
  archivePrefix = {arXiv},
  primaryClass  = {cs.RO},
  url           = {https://arxiv.org/abs/2602.12684}
}

@inproceedings{liu2025rdt,
  title     = {{RDT-1B}: A Diffusion Foundation Model for Bimanual Manipulation},
  author    = {Liu, Songming and Wu, Lingxuan and Li, Bangguo and Tan, Hengkai and Chen, Huayu and Wang, Zhengyi and Xu, Ke and Su, Hang and Zhu, Jun},
  booktitle = {International Conference on Learning Representations},
  year      = {2025},
  eprint    = {2410.07864},
  archivePrefix = {arXiv},
  url       = {https://arxiv.org/abs/2410.07864}
}

@misc{pertsch2025fast,
  title         = {{FAST}: Efficient Action Tokenization for Vision-Language-Action Models},
  author        = {Pertsch, Karl and Stachowicz, Kyle and Ichter, Brian and Driess, Danny and Nair, Suraj and Vuong, Quan and Mees, Oier and Finn, Chelsea and Levine, Sergey},
  year          = {2025},
  eprint        = {2501.09747},
  archivePrefix = {arXiv},
  primaryClass  = {cs.RO},
  url           = {https://arxiv.org/abs/2501.09747}
}

@inproceedings{liu2023libero,
  title     = {{LIBERO}: Benchmarking Knowledge Transfer for Lifelong Robot Learning},
  author    = {Liu, Bo and Zhu, Yifeng and Gao, Chongkai and Feng, Yihao and Liu, Qiang and Zhu, Yuke and Stone, Peter},
  booktitle = {Advances in Neural Information Processing Systems, Datasets and Benchmarks Track},
  year      = {2023},
  eprint    = {2306.03310},
  archivePrefix = {arXiv},
  url       = {https://arxiv.org/abs/2306.03310}
}

@misc{mees2021calvin,
  title         = {{CALVIN}: A Benchmark for Language-Conditioned Policy Learning for Long-Horizon Robot Manipulation Tasks},
  author        = {Mees, Oier and Hermann, Lukas and Rosete-Beas, Erick and Burgard, Wolfram},
  year          = {2021},
  eprint        = {2112.03227},
  archivePrefix = {arXiv},
  primaryClass  = {cs.RO},
  url           = {https://arxiv.org/abs/2112.03227}
}

@inproceedings{li2025simpler,
  title     = {Evaluating Real-World Robot Manipulation Policies in Simulation},
  author    = {Li, Xuanlin and Hsu, Kyle and Gu, Jiayuan and Pertsch, Karl and Mees, Oier and Walke, Homer Rich and Fu, Chuyuan and Lunawat, Ishikaa and Sieh, Isabel and Kirmani, Sean and Levine, Sergey and Wu, Jiajun and Finn, Chelsea and Su, Hao and Vuong, Quan and Xiao, Ted},
  booktitle = {Proceedings of the Conference on Robot Learning},
  series    = {Proceedings of Machine Learning Research},
  volume    = {270},
  year      = {2025},
  url       = {https://proceedings.mlr.press/v270/li25c.html},
  eprint    = {2405.05941},
  archivePrefix = {arXiv}
}

@misc{zhou2025liberopro,
  title         = {{LIBERO-PRO}: Towards Robust and Fair Evaluation of Vision-Language-Action Models Beyond Memorization},
  author        = {Zhou, Xueyang and Xu, Yangming and Tie, Guiyao and Chen, Yongchao and Zhang, Guowen and Chu, Duanfeng and Zhou, Pan and Sun, Lichao},
  year          = {2025},
  eprint        = {2510.03827},
  archivePrefix = {arXiv},
  primaryClass  = {cs.CV},
  url           = {https://arxiv.org/abs/2510.03827}
}

@inproceedings{wang2026liberox,
  title     = {{LIBERO-X}: Robustness Litmus for Vision-Language-Action Models},
  author    = {Wang, Guodong and Zhang, Chenkai and Liu, Qingjie and Zhang, Jinjin and Cai, Jiancheng and Liu, Junjie and Liu, Xinmin},
  booktitle = {Proceedings of Robotics: Science and Systems},
  year      = {2026},
  url       = {https://roboticsconference.org/program/papers/97/},
  eprint        = {2602.06556},
  archivePrefix = {arXiv},
  primaryClass  = {cs.CV}
}

@misc{kim2026liberopara,
  title         = {{LIBERO-Para}: A Diagnostic Benchmark and Metrics for Paraphrase Robustness in {VLA} Models},
  author        = {Kim, Chanyoung and Kim, Minwoo and Kang, Minseok and Kim, Hyunwoo and Jung, Dahuin},
  year          = {2026},
  eprint        = {2603.28301},
  archivePrefix = {arXiv},
  primaryClass  = {cs.RO},
  url           = {https://arxiv.org/abs/2603.28301}
}

@misc{hou2026langgap,
  title         = {{LangGap}: Diagnosing and Closing the Language Gap in Vision-Language-Action Models},
  author        = {Hou, Yuchen and Zhao, Lin},
  year          = {2026},
  eprint        = {2603.00592},
  archivePrefix = {arXiv},
  primaryClass  = {cs.RO},
  url           = {https://arxiv.org/abs/2603.00592}
}

@misc{guruprasad2025multinet,
  title         = {An Open-Source Software Toolkit \& Benchmark Suite for the Evaluation and Adaptation of Multimodal Action Models},
  author        = {Guruprasad, Pranav and Wang, Yangyue and Chowdhury, Sudipta and Song, Jaewoo and Sikka, Harshvardhan},
  year          = {2025},
  eprint        = {2506.09172},
  archivePrefix = {arXiv},
  primaryClass  = {cs.LG},
  url           = {https://arxiv.org/abs/2506.09172},
  note          = {Accepted at the ICML 2025 CODEML Workshop}
}

@misc{choi2026vlaeval,
  title         = {{vla-eval}: A Unified Evaluation Harness for Vision-Language-Action Models},
  author        = {Choi, Suhwan and Lee, Yunsung and Park, Yubeen and Kim, Chris Dongjoo and Krishna, Ranjay and Fox, Dieter and Yu, Youngjae},
  year          = {2026},
  eprint        = {2603.13966},
  archivePrefix = {arXiv},
  primaryClass  = {cs.RO},
  url           = {https://arxiv.org/abs/2603.13966}
}

@misc{cadene2026lerobot,
  title         = {{LeRobot}: An Open-Source Library for End-to-End Robot Learning},
  author        = {Cadene, Remi and Aliberts, Simon and Capuano, Francesco and Aractingi, Michel and Zouitine, Adil and Kooijmans, Pepijn and Choghari, Jade and Russi, Martino and Pascal, Caroline and Palma, Steven and Shukor, Mustafa and Moss, Jess and Soare, Alexander and Aubakirova, Dana and Lhoest, Quentin and Gallouedec, Quentin and Wolf, Thomas},
  year          = {2026},
  eprint        = {2602.22818},
  archivePrefix = {arXiv},
  primaryClass  = {cs.RO},
  url           = {https://arxiv.org/abs/2602.22818}
}

@misc{chen2026hazardarena,
  title         = {{HazardArena}: Evaluating Semantic Safety in Vision-Language-Action Models},
  author        = {Chen, Zixing and Gao, Yifeng and Wang, Li and Zhao, Yunhan and Liu, Yi and Li, Jiayu and Zheng, Xiang and Wu, Zuxuan and Wang, Cong and Ma, Xingjun and Jiang, Yu-Gang},
  year          = {2026},
  eprint        = {2604.12447},
  archivePrefix = {arXiv},
  primaryClass  = {cs.RO},
  url           = {https://arxiv.org/abs/2604.12447}
}

@inproceedings{fei2026liberoplus,
  title     = {{LIBERO-Plus}: A Progressive Robustness Benchmark for Visual-Language-Action Models},
  author    = {Fei, Senyu and Wang, Siyin and Shi, Junhao and Dai, Zihao and Cai, Jikun and Qian, Pengfang and Ji, Li and He, Xinzhe and Zhang, Shiduo and Fei, Zhaoye and Fu, Jinlan and Gong, Jingjing and Qiu, Xipeng},
  booktitle = {Proceedings of the IEEE/CVF Conference on Computer Vision and Pattern Recognition},
  pages     = {38574--38583},
  month     = {June},
  year      = {2026}
}

@article{fastwam,
  title         = {{Fast-WAM}: Do World Action Models Need Test-time Future Imagination?},
  author        = {Yuan, Tianyuan and Dong, Zibin and Liu, Yicheng and Zhao, Hang},
  journal       = {arXiv preprint arXiv:2603.16666},
  year          = {2026},
  eprint        = {2603.16666},
  archivePrefix = {arXiv},
  primaryClass  = {cs.CV},
  url           = {https://arxiv.org/abs/2603.16666}
}

@misc{nvidia2026groot17,
  author       = {{NVIDIA}},
  title        = {{NVIDIA Isaac GR00T N1.7}: A Foundation Model for Generalist Robots},
  year         = {2026},
  howpublished = {GitHub repository and model release},
  url          = {https://github.com/NVIDIA/Isaac-GR00T/releases/tag/n1.7-release},
  note         = {N1.7 release, accessed July 20, 2026}
}

@misc{yuan2026fastwam,
  title         = {{Fast-WAM}: Do World Action Models Need Test-time Future Imagination?},
  author        = {Yuan, Tianyuan and Dong, Zibin and Liu, Yicheng and Zhao, Hang},
  year          = {2026},
  eprint        = {2603.16666},
  archivePrefix = {arXiv},
  primaryClass  = {cs.RO},
  url           = {https://arxiv.org/abs/2603.16666}
}

@misc{kim2026cosmospolicy,
  title         = {Cosmos Policy: Fine-Tuning Video Models for Visuomotor Control and Planning},
  author        = {Kim, Moo Jin and Gao, Yihuai and Lin, Tsung-Yi and Lin, Yen-Chen and Ge, Yunhao and Lam, Grace and Liang, Percy and Song, Shuran and Liu, Ming-Yu and Finn, Chelsea and Gu, Jinwei},
  year          = {2026},
  eprint        = {2601.16163},
  archivePrefix = {arXiv},
  primaryClass  = {cs.RO},
  url           = {https://arxiv.org/abs/2601.16163}
}

@misc{zhang2026wamrobustness,
  title         = {Do World Action Models Generalize Better than {VLA}s? A Robustness Study},
  author        = {Zhang, Zhanguang and Li, Zhiyuan and Rahmati, Behnam and Yang, Rui Heng and Ma, Yintao and Rasouli, Amir and Pakdamansavoji, Sajjad and Wu, Yangzheng and Zhang, Lingfeng and Cao, Tongtong and Wen, Feng and Wang, Xinyu and Quan, Xingyue and Zhang, Yingxue},
  year          = {2026},
  eprint        = {2603.22078},
  archivePrefix = {arXiv},
  primaryClass  = {cs.RO},
  url           = {https://arxiv.org/abs/2603.22078}
}

@misc{zhang2025vlaarena,
  title={VLA-Arena: An Open-Source Framework for Benchmarking Vision-Language-Action Models},
  author={Borong Zhang and Jiahao Li and Jiachen Shen and Yishuai Cai and Yuhao Zhang and Yuanpei Chen and Juntao Dai and Jiaming Ji and Yaodong Yang},
  year={2025},
  eprint={2512.22539},
  archivePrefix={arXiv},
  primaryClass={cs.RO},
  url={https://arxiv.org/abs/2512.22539}
}

\clearpage
\appendix

\section{Unified Execution View and Failure Taxonomy}
\label{app:execution-view}

Formally, we view a VLA evaluation episode as a trajectory over instruction-conditioned observations, actions, state transitions, and outcomes. Given a language instruction $I$, an observation stream $O_t$ containing visual inputs and robot state, and a policy or inference procedure $\pi$, the model produces an action $A_t = \pi(I, O_{\leq t})$. Executing $A_t$ induces an environment transition to $S_{t+1}$ and eventually yields an outcome $Y$, such as success, failure, or partial completion.

IndustrialVLA-Bench uses an execution-oriented failure vocabulary to support qualitative interpretation of diagnostic results. We distinguish seven broad failure categories. \textbf{Instruction errors} occur when the model ignores, misinterprets, or inconsistently follows task instructions, including paraphrases, ordering constraints, negations, or object attributes. \textbf{Perception errors} occur when the model or observation pipeline fails to recognize task-relevant objects, visual states, or scene changes. \textbf{Grounding errors} occur when the model recognizes relevant entities but associates the instruction with the wrong object, location, spatial relation, or affordance. \textbf{Planning errors} occur when the model selects an inappropriate subgoal or action sequence despite having sufficient perceptual and instructional information. \textbf{Execution errors} occur when an otherwise reasonable action intention is not translated into successful low-level control. \textbf{Recovery errors} occur when the model fails to correct or recover from an intermediate mistake or unexpected state transition. \textbf{Runtime/reproduction errors} cover failures caused by missing checkpoints, unsupported wrappers, dependency conflicts, server--client failures, action-normalization mismatches, or unresolved evaluation-path differences. This vocabulary is used as an interpretive framework for discussing representative failure modes rather than as a fully annotated causal label assigned to every failed episode.

Finally, IndustrialVLA-Bench defines an evidence taxonomy to prevent unsupported or partially comparable results from being merged into the main leaderboard. \textbf{Main evidence} consists of directly comparable official-checkpoint results produced through official or protocol-faithful evaluation paths. These results use the official task definitions, success rules, trial counts, action interfaces, and aggregation protocols, and are eligible for the main leaderboard. \textbf{Secondary evidence} consists of useful but partially comparable results, such as results on CALVIN or SimplerEnv when only a subset of models has official support or when benchmark coverage is uneven \cite{mees2021calvin,li2025simpler}. \textbf{Diagnostic evidence} consists of stress-test results that probe robustness, language sensitivity, generalization, physical perturbation, or safety behavior, and should be interpreted as axis-specific evidence rather than as a replacement for the main leaderboard \cite{zhou2025liberopro,wang2026liberox,kim2026liberopara,hou2026langgap,chen2026hazardarena}. \textbf{Appendix evidence} consists of incomplete, unsupported, wrapper-dependent, near-reproduction, or reduced-trial runs that may be useful for transparency but are not eligible for direct ranking. Results based on hand-written paraphrases, modified success rules, altered benchmark definitions, or synthetic sanity checks are recorded only as sanity-check evidence.

This evidence taxonomy reflects a core principle of IndustrialVLA-Bench: comparability must be earned rather than assumed. Because open-source VLA models differ substantially in benchmark support, runtime architecture, checkpoint availability, and engineering maturity, forcing all available numbers into a single table would create misleading comparisons. Instead, our benchmark explicitly records the evidence level of each result, the protocol used to obtain it, and the reason why it is eligible or ineligible for main-table promotion. This allows IndustrialVLA-Bench to support broad model coverage while preserving a strict standard for direct comparison.

\section{Evidence Policy and Comparability Gates}
\label{app:evidence}

\subsection{Model Inclusion Criteria}
\label{app:inclusion}

At the July 2026 snapshot, a model was considered when (i) a public checkpoint and implementation were available; (ii) it supported instruction-conditioned manipulation or embodied action generation; (iii) its observation and action interfaces could be mapped to LIBERO without changing task semantics; (iv) its inference configuration could be documented; and (v) it did not require a closed remote API. The final six-model set additionally reflects the practical goal of coverage across all three tracks. This is a purposive inclusion process, not a systematic enumeration of all eligible public releases; the paper therefore makes no coverage claim beyond the named checkpoints.

\subsection{Evidence Views}
\label{sec:evidence-views}

IndustrialVLA-Bench separates reported results according to their experimental role and comparability status rather than merging all available numbers into a single leaderboard.

\textbf{Main capability evidence.}
Clean LIBERO results obtained using public checkpoints and protocol-faithful evaluation paths constitute strict capability evidence. The clean table also shows NR and PV rows for transparency, but those rows are not evidence-equivalent and are excluded from strict rank claims.

\textbf{Diagnostic evidence.}
LIBERO-Plus robustness results and LIBERO-Para language-sensitivity results constitute diagnostic evidence. They are reported separately because controlled non-language perturbations and semantics-preserving instruction paraphrases evaluate different model properties and should not be collapsed into a single robustness score.

\textbf{Deployability and reproducibility evidence.}
Execution and reproduction evidence consists of observed latency, peak VRAM, parameter count, diffusion configuration, runtime mode, setup burden, checkpoint provenance, wrapper deviations, and verification status. Because hardware and policy-call boundaries are not normalized in the current release, latency and memory characterize observed runs rather than a controlled cross-model efficiency ranking.

\textbf{Appendix or unresolved evidence.}
Partial, reduced-trial, estimated, near-reproduction, unsupported, or otherwise non-comparable runs are retained as appendix or unresolved evidence. Such results may be useful for transparency and debugging, but they are not eligible for strict leaderboard claims.

This separation reflects a core principle of IndustrialVLA-Bench: comparability must be demonstrated rather than assumed. Each reported result is accompanied by its execution path, protocol status, and evidence strength, preventing partially comparable or unresolved runs from being interpreted as equally strong benchmark evidence.

\subsection{Evaluation Track}
\label{sec:evaluation-track}

IndustrialVLA-Bench uses an \textbf{official-checkpoint evaluation track}. It evaluates publicly released checkpoints through official or documented inference paths, asking how the named releases behave out of the box and how strong the supporting execution evidence is. NR and PV paths remain visible but are not described as protocol-faithful.

For every model, the checkpoint and inference configuration are held fixed across evaluation seeds and across LIBERO, LIBERO-Plus, and LIBERO-Para. No model is adapted or fine-tuned on LIBERO-Plus or LIBERO-Para. In particular, LIBERO-Plus evaluation uses the same clean-task checkpoint without training on the evaluated perturbations, while LIBERO-Para evaluation uses the same checkpoint without training on paraphrased instructions.

The current release does not claim experimental coverage of CALVIN, SimplerEnv, LIBERO-PRO, LIBERO-X, LangGap, AGNOSTOS, VLA-Arena, real-robot safety evaluation, or broader cross-benchmark generalization. These settings may provide useful future extensions, but they are not part of the present experimental claims.

An equal-budget adaptation comparison is outside the scope of the current study because it would require retraining heterogeneous systems under a shared data, optimization, and compute budget. We leave such an analysis to future work and do not mix adaptation claims with released-checkpoint performance.

\subsection{Comparability and Ranking Policy}
\label{sec:ranking-policy}

A result is eligible for the main leaderboard only if it passes the following comparability gates: the same benchmark version or a documented compatible version; the same task suites and task definitions; a public official checkpoint or documented checkpoint provenance; preserved observation and action semantics, including action normalization; official success rules, trial counts, episode termination, and aggregation procedures; no hidden prompt, wrapper, control, or reset advantage; and disclosed wrapper changes limited to protocol-faithful engineering functions such as logging, path forwarding, device placement, or deployment measurement.

IndustrialVLA-Bench reports results through separate evidence views. The \textbf{main leaderboard} reports clean LIBERO capability under directly comparable official-checkpoint conditions. The \textbf{diagnostic tables} report LIBERO-Plus robustness and LIBERO-Para language-sensitivity results separately. The \textbf{deployability table} reports inference latency, peak VRAM, runtime mode, setup burden, and reproduction status. The \textbf{model profile} interprets these axes jointly, but IndustrialVLA-Bench does not aggregate them into a universal scalar score.

This policy reflects the central claim of IndustrialVLA-Bench: it does not ask which model is universally best; it asks which model is best under which evidence axis. By separating evidence types and enforcing comparability gates, IndustrialVLA-Bench provides a fairer basis for selecting, reproducing, and improving open-source industrial VLA systems.

\section{Benchmark Harness and Reproducibility}
\label{sec:harness-reproducibility}

A benchmark for open-source industrial VLA systems must be reproducible not only at the level of final numbers, but also at the level of execution paths, checkpoints, environments, commands, logs, and result promotion decisions. IndustrialVLA-Bench therefore includes a benchmark harness that standardizes how results are launched, recorded, normalized, validated, and promoted into paper tables. The purpose of the harness is not to replace official repositories with a new implementation, but to make official and protocol-faithful executions traceable and comparable.

\subsection{Official-First Execution}
\label{sec:official-first-execution}

IndustrialVLA-Bench follows an official-first execution principle. For each model, we prioritize the official repository, official checkpoint, official inference path, and official evaluation script whenever they are available. A wrapper is allowed only when it preserves benchmark semantics and is limited to engineering functions such as path forwarding, logging, environment isolation, device placement, result formatting, and deployment measurement. It must not modify task definitions, success rules, prompt semantics, observation contents, action normalization, control frequency, reset behavior, or episode termination.

This principle is critical for VLA evaluation. Unlike static prediction tasks, VLA scores can be affected by small execution-level differences. A seemingly minor wrapper change may alter camera observations, action timing, prompt formatting, control interfaces, or termination behavior, making success rates non-comparable. IndustrialVLA-Bench therefore treats the evaluation path itself as part of the benchmark artifact. Each result must disclose whether it is produced by an official script, a protocol-faithful wrapper, a near reproduction, or an unsupported custom path. Only official or protocol-faithful executions are eligible for main-leaderboard promotion.

\subsection{Environment Isolation}
\label{sec:environment-isolation}

Open-source VLA repositories often depend on different versions of Python, CUDA, PyTorch, MuJoCo, robosuite, transformers, flash-attn, diffusion libraries, simulation backends, and robot-control utilities. Forcing all models into a single unified software environment can break official execution paths or introduce unofficial engineering changes. IndustrialVLA-Bench therefore isolates environments at the model level. Each model is evaluated in its own conda, Docker, or documented runtime environment, following the official installation instructions as closely as possible.

For every run, the harness records package versions, CUDA version, GPU type, GPU memory, CPU type when available, operating system, driver version, repository commit, checkpoint identifier, and benchmark version. Environment isolation is not treated as a nuisance but as a deployability signal. If a model cannot be installed, requires undocumented dependencies, conflicts with required benchmark packages, or fails under its official instructions, the failure is recorded as reproducibility evidence rather than silently discarded. This allows IndustrialVLA-Bench to report not only which models achieve high success rates, but also which models can be reliably reproduced by downstream users.

\subsection{Command Manifests and Normalized Results}
\label{sec:command-manifest}

IndustrialVLA-Bench uses command manifests to make every reported number traceable to an executable run. A manifest records the model name, repository URL and commit, checkpoint identifier, benchmark suite, task split, entry point, launch command, seed, trial count, hardware configuration, environment name, raw log path, output directory, success metric, latency measurement, VRAM measurement, runtime mode, and run status. A simplified schema is shown below:

\begin{center}
\small
\begin{tabular}{ll}
\textbf{Field} & \textbf{Description} \\
\midrule
model & Model or model family name \\
repo\_commit & Repository commit or release tag \\
checkpoint\_id & Checkpoint name, path, or hash \\
benchmark & Benchmark suite and version \\
task\_suite & Task split or evaluation subset \\
entry\_point & Official script or protocol-faithful wrapper \\
command & Full launch command \\
seed & Random seed when applicable \\
num\_trials & Number of evaluation episodes \\
raw\_log & Path to raw execution logs \\
success\_rate & Aggregated success metric \\
latency & Latency per action step or episode-level timing \\
peak\_vram & Peak GPU memory usage \\
runtime\_mode & Local, Docker, server-client, or remote-like mode \\
status & Complete, partial, failed, unresolved, or appendix-only \\
\bottomrule
\end{tabular}
\end{center}

The normalized result schema contains one row per model--benchmark--suite--track configuration, with fields for success rate, run-level deviation, episode count, evidence tier, comparability status, runtime statistics, reproduction status, and wrapper notes. A row is promoted only after its normalized record is matched to the available logs and manifest; otherwise it remains NR or PV. This design preserves failed and partial paths rather than silently deleting them.

\subsection{Result Promotion Checklist}
\label{sec:promotion-checklist}

IndustrialVLA-Bench uses a result promotion checklist to decide whether a run can enter the main paper tables. A result is eligible for the main leaderboard only if the raw log exists, the repository commit and checkpoint provenance are recorded, the benchmark version and task suite are documented, the metric aggregation follows the official definition, the trial count matches the protocol, wrapper deviations are disclosed, and observation/action semantics are preserved. Deployment metadata, including latency, peak VRAM, runtime mode, and reproduction status, must also be recorded when the run is complete.

Failed, partial, unsupported, or unresolved runs are not deleted. Instead, they are assigned an explicit status and reported as appendix, diagnostic, or reproducibility evidence when informative. This policy prevents the benchmark from reporting only successful reproductions while hiding engineering failures. It also prevents non-comparable numbers from entering the main leaderboard. For example, a run with reduced trials, modified success rules, hand-written paraphrases, altered reset behavior, unofficial action wrappers, or unresolved checkpoint mismatch may still be useful for debugging or transparency, but it is not promoted to main evidence.

The checklist protects comparability and traceability. A PF result must connect its number to a concrete command, environment, checkpoint, log, and aggregation rule. Rows that do not yet satisfy that chain remain visibly NR or PV.

\section{Extended Experimental Setup}
\label{app:setup}

\paragraph{Checkpoints and evaluation paths.} We prioritize officially released checkpoints and official evaluation implementations whenever they are available. When an official end-to-end path is unavailable, we construct a protocol-faithful reproduction that preserves the benchmark task definitions, observation and action semantics, success conditions, episode horizon, and aggregation rules. Any unavoidable deviation in model wrappers, prompt templates, image preprocessing, or runtime configuration is disclosed.

\paragraph{Random seeds and aggregation.} Every reported model--benchmark configuration is evaluated through three complete runs with fixed run-level seeds $\{s_1,s_2,s_3\}=\{1,7,42\}$. A run seed initializes the simulator reset and task-order random-number generators, and the policy-sampling generator when stochastic sampling is enabled; it is neither an episode index nor a training seed. The checkpoint and inference configuration are held fixed across the three runs, and each run evaluates the complete official task set. Standard LIBERO uses 50 trials per task, giving 2{,}000 episodes per seed and 6{,}000 episodes over three seeds. LIBERO-Plus and LIBERO-Para use one episode per official task configuration, giving 10{,}030 and 4{,}092 episodes per seed, respectively. We report the run-level mean and population standard deviation
\[
\bar r=\frac{1}{3}\sum_{i=1}^{3}r_i,
\qquad
\mathrm{Std}=\sqrt{\frac{1}{3}\sum_{i=1}^{3}(r_i-\bar r)^2},
\]
where $r_i$ is the success rate of run $i$. ``Std'' therefore describes variation across the three reported runs rather than an episode-level binomial error bar. %
The three seeds share a fixed evaluation set: standard LIBERO uses the
official initial-state files, and LIBERO-Plus and LIBERO-Para evaluate one
episode per official task configuration. The seeds therefore vary simulator
and policy-sampling nondeterminism over a fixed configuration set, not the
configuration sample itself. The reported deviations describe replay
variation under that fixed set. They are not sampling error bars over task
configurations and must not be read as confidence intervals. Quantifying
configuration-level uncertainty would require bootstrap resampling over
task configurations, which we leave to the next release.

\paragraph{Reported metrics.} For clean LIBERO evaluation, we report the success rate for each suite and the unweighted average over the four suites. For LIBERO-Plus, we report success under each perturbation dimension and an unweighted robustness average over the six non-linguistic dimensions. For LIBERO-Para, we report success under instruction variations separately rather than mixing language sensitivity with visual robustness.

\paragraph{Runtime measurements.} Peak VRAM is measured during inference after model loading and warm-up. Latency is the mean wall-clock time per policy invocation and excludes environment reset. One invocation may emit a model-specific chunk of $k$ actions, so the reported value is not amortized per executed action. Model-specific hardware, precision, batching, preprocessing, communication, and chunking prevent a controlled cross-model latency rank. A future normalized release should report the same device and precision, $k$, per-call and amortized latency, effective control frequency, and median/P90/P95 statistics.

\paragraph{Evidence status.} \emph{Protocol-faithful (PF)} indicates a reproduced path that preserves benchmark semantics and aggregation rules. \emph{Near-reproduction (NR)} indicates disclosed wrapper- or setup-dependent deviations. \emph{Pending verification (PV)} indicates that the path has not yet passed the promotion checklist. The current audit assigns PF to OpenPI, GR00T-N1.7, and FastWAM; NR to UnifoLM-VLA-0; and PV to Xiaomi-Robotics-0 and Cosmos Policy. Status is orthogonal to the numerical score.

\end{document}